\documentclass[letterpaper, 10 pt, conference]{ieeeconf}  

\usepackage{graphicx} 
\usepackage{subfigure} 
\usepackage{wrapfig}
\usepackage{color, colortbl}

\usepackage{cite}

\usepackage{algorithm}
\usepackage{algorithmic}
\usepackage[table]{xcolor}

\newcommand{\rd}{{\mathrm d}}

\newcommand{\vx}{{\bf x}}

\usepackage{amsmath} 
\usepackage{amssymb}  

\usepackage{xcolor}

\IEEEoverridecommandlockouts                              

\title{\LARGE \bf Vision-Based High Speed Driving with a Deep Dynamic Observer 
   }

\author{Paul Drews, Grady Williams, Brian Goldfain, Evangelos A. Theodorou, and James M. Rehg 
\thanks{The authors are with Institute for Robotics and Intelligent Machines (IRIM) at the Georgia Institute of Technology, 
Atlanta, 
GA, USA. 
Email: pdrews3@gatech.edu}
}
\begin{document}

\maketitle
\thispagestyle{empty}
\pagestyle{empty}

\global\csname @topnum\endcsname 0
\global\csname @botnum\endcsname 0

\begin{abstract} 
In this paper we present a framework for combining deep learning-based road detection, particle filters, and Model Predictive Control (MPC) to drive aggressively using only a monocular camera, IMU, and wheel speed sensors. This framework uses deep convolutional neural networks combined with LSTMs to learn a local cost map representation of the track in front of the vehicle. A particle filter uses this dynamic observation model to localize in a schematic map, and MPC is used to drive aggressively using this particle filter based state estimate.  We show extensive real world testing results, and demonstrate reliable operation of the vehicle at the friction limits on a complex dirt track.  We reach speeds above 27 mph (12 m/s) on a dirt track with a 105 foot (32m) long straight using our 1:5 scale test vehicle.
\end{abstract}



\section{Introduction}

\begin{figure}[t]
	\begin{center}
    \includegraphics[width=0.95\columnwidth]{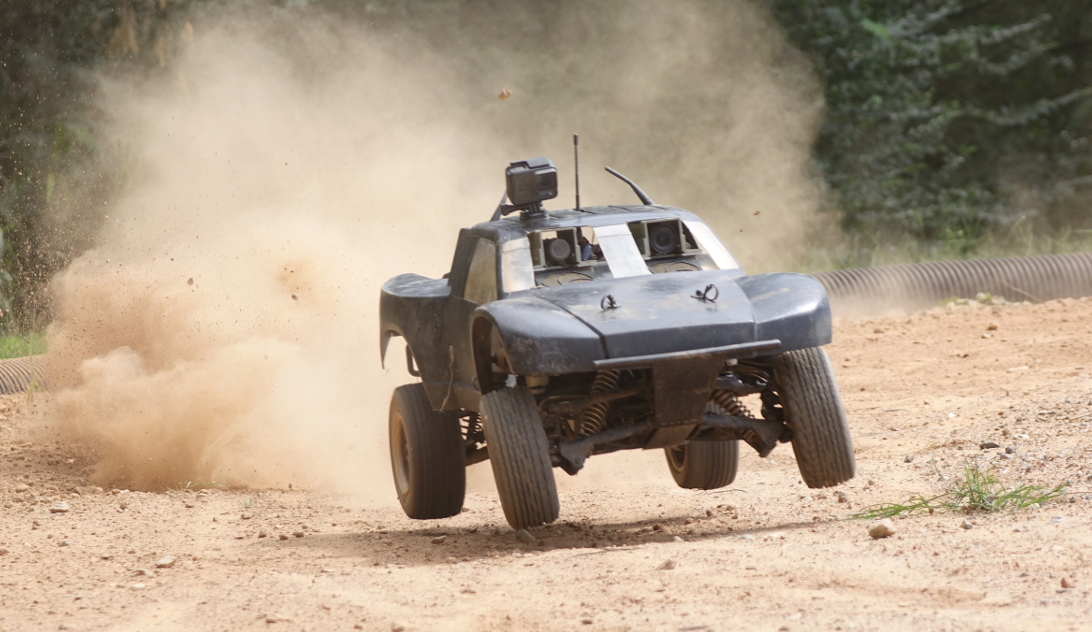}
    \caption{AutoRally vehicle navigating a bump on the track at high speed during testing}
    \label{fig:aggressive_hop}
    \end{center}
    \vskip -0.2in
\end{figure}

\begin{figure*}
\begin{center}
{
\includegraphics[width=2\columnwidth]{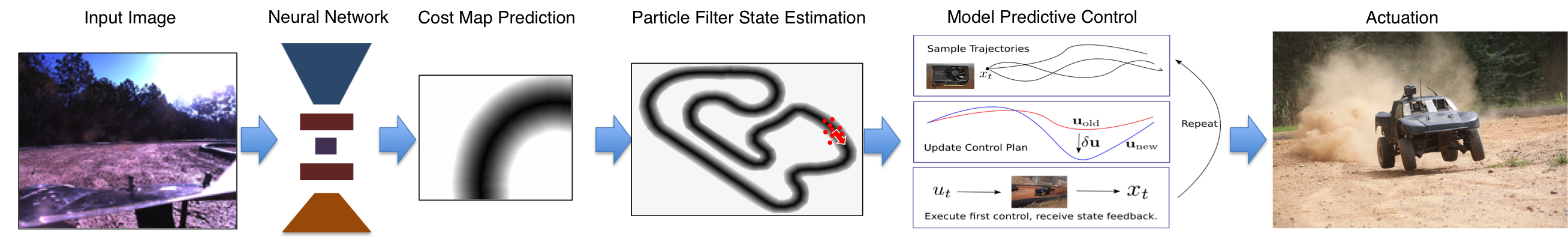}
}
\caption{Full system diagram.}
\label{fig:systemDiagram}
\end{center}
\vskip -0.2in
\end{figure*} 

\begin{figure}[b]
	\begin{center}
	\includegraphics[width=0.8\columnwidth]{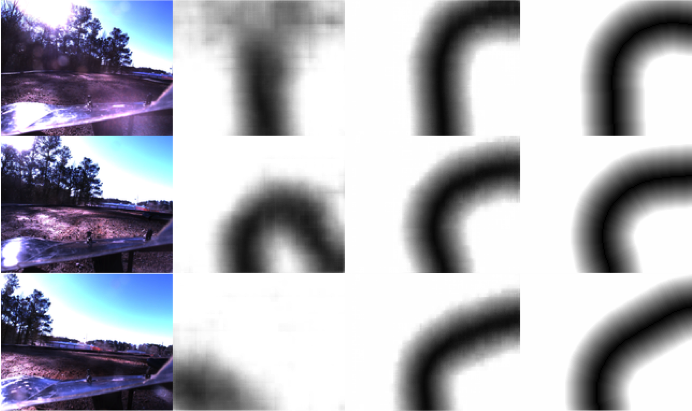}
	\caption{Failure case for single frame network.  From left to right: input image, single frame network output, lstm output, ground truth cost map}
    \label{fig:lstmsequence}
    \end{center}
\end{figure}




We address the challenging task of \emph{aggressive driving}--- autonomous control of a ground vehicle operating close to the limits of handling, often with high sideslip angles, as is required in rally racing  (and possibly for collision avoidance). 
We build on our recent work on aggressive driving~\cite{williams2017information,williams2016}, using the open source 1:5 scale \emph{AutoRally} platform described in~\cite{goldfain2018autorally}.
Much of our prior work relies on high-quality GPS for global position estimation and localization relative to a known map, which limits its applicability by requiring expensive sensors and excluding GPS-denied areas.
An exception is~\cite{drews17aggressive}, which presents a vision-based (i.e. non-GPS) driving solution based on regressing a local cost map from monocular camera images which is then used directly for MPC-based control.
%
%
%
%

However, treating each input frame separately leads to a very challenging learning problem.
This stems from the limited field of view and low vantage point of a camera mounted on a ground vehicle, which makes it difficult to generate cost maps that extend sufficiently far in front of the vehicle, given the high speed of travel.  Our solution to the limitations of single frame cost map prediction is to incorporate recurrence in the form of an LSTM model, making it possible to exploit the temporal continuity of the track via the structure of the on-board camera video. Fig.~\ref{fig:lstmsequence} illustrates both the difficulty of predicting a high quality cost map from a single image under challenging conditions, and the significant improvement that arises from incorporating recurrence into the model. The single-frame cost map predictions shown in the second column are significantly less accurate than the cost maps produced by the LSTM model (third column), which is described in Section~\ref{sec:cnn}. 



Our second innovation is the development of a particle filter estimator for the vehicle state which treats the generated cost maps as a sequence of measurements. This is in contrast to directly controlling the vehicle from the local cost maps, as was done in~\cite{drews17aggressive}. Instead, our architecture incorporates two dynamic measurement processes: an LSTM that processes on-board video to estimate the evolving cost map, and a particle filter state estimator which drives the model predictive control algorithm. A benefit of this approach is that the vision-derived cost maps can be combined in a principled way with other sensors, such as an IMU and wheel speed encoders. This approach also naturally decouples the state estimator and controller, which can leverage mature, well-understood technologies such as particle filters and MPC, from the video-based deep neural network model (i.e. LSTM) which provides "black box" estimates of the cost map. Thus the vehicle state provides an interpretable and useful latent representation which is helpful in diagnosing issues with the cost map predictor. This is in contrast to standard end-to-end approaches to learning control~\cite{Pan-RSS-18}, which typically lack such informative intermediate latent representations. 


In summary, this paper introduces an alternative approach to autonomous high-speed driving in which a local cost map generator in the form of a video-based LSTM is used as the measurement process for a particle filter state estimator. This allows us to obtain a global position estimate against a schematic map without the use of GPS. Moreover, the recurrent visual models are able to take advantage of the temporal structure of on-board video, improving the accuracy of cost map prediction.  We make the following contributions:
\begin{itemize}
  \item A novel encoder-decoder deep learning architecture utilizing an LSTM which improves performance over single frame networks by incorporating temporal information.
  \item A GPU-based particle filter state estimator which can take the output of this dynamic observation model and smooth it using IMU and wheel speeds.  This allows localization in real time against a schematic track map.
  \item Repeatable performance at the limits of the system's capability using only monocular cameras for 
localization.  We are able to repeatably beat the best single lap performed by an experienced human test driver who provided all of the system identification data.
\end{itemize}

\section{Related Work}



Several approaches have been taken to the problem of aggressive autonomous driving.  In \cite{funke2012limits}, an analytic approach is explored.  The performance limits of a vehicle are pushed using a simple model-based feedback controller and extensive pre-planning to follow a racing line around a track. More recently, \cite{keivan2013realtime} showed the benefits of model predictive control on a 1:10 scale vehicle following waypoints through a challenging obstacle course.  However, these approaches all rely on highly accurate position from an external source such as GPS or motion capture.  

Portions of the solution that we propose have been studied in isolation.  There is a great deal of literature about localizing from camera images.  One successful approach is found in the Simultaneous Localization And Mapping (SLAM) literature.  Methods such \cite{engel2014lsd,mur2015orb} use whole image matching and keypoints while infering and leveraging the 3d structure of the scene.  These systems typically provide position relative to a generated map and could be suitable for our use case.  However, these methods fall short in close to the ground, monocular, high dynamic and vibration video such as we have.  Another approach that doesn't require explicit estimation of 3d scene geometry is to directly regress the position from camera images.  Works such as \cite{hays2008im2gps} and more recently \cite{kendall2015modelling} show promise.  However, they do not have the fine-grained accuracy that we require in order to drive aggressively in a tightly constrained track.  Semantic segmentation methods such as \cite{arbelaez2012semantic} and \cite{long2015fully} may be used to obtain drivability regions in the image as our method does. However, these tend to be computationally heavy and require a planar projection or depth data to transform to the cost map representation we use.


Autonomous driving and control as an end-to-end learning problem is an active area of research. Prior work on deep learning and model predictive control includes end-to-end methods \cite{bojarski2016end,Pan-RSS-18,Zhang2016} and encoder-decoder architectures \cite{Lenz2015DeepMPCLD,WAHLSTROM20151059} to perform predictive control using raw observations. In  \cite{Pan-RSS-18} an MPC controller is used as a teacher to train  a Convolutional Neural Network that maps raw observations such as wheel speed, acceleration, and images directly to throttle and steering. 
While it elegantly encompasses the our entire system as a single learning problem, it has severe drawbacks. When errors occur, there is very little ability to probe the reasons for the failure.  Because this solution is not modular, it must be re-learned if any part of the system changes.  Because of the modularity of our system, if for example the vehicle changes we can simply swap out the dynamics model.  End-to-end immitation learning methods are also inherently limited by their teacher.  However, our system is able handily outperform the best lap from the training set.

Some of the most related work comes in direct affordance based control.  In \cite{schenck2017visual}, a recurrent learning system is used to estimate liquid volumes as an affordance from video.  However, they learn most of the required dynamics, including state propagation, and control with a simple PID controller.  Our system has much more stringent control requirements, necessitating more accurate state estimate.  In \cite{chen2015deepdriving}, lane affordances are directly learned and fed to a simple lane tracking controller.  However, this work does not address the aggressive driving regime.  

Additional closely related work shows the integration of particle filtering and vision based observations in \cite{miller2008particle}.  More recent work such as \cite{rose2014integrated} show more modern results other filter methods.  They show good performance with the addition of lane markings, but do not utilize learning to help with the vision portion of the problem and do not address the particular problems of the aggressive driving regime.
\section{Deep Neural Networks for Localization}

At the highest level, planning often takes place on a schematic map. We define a schematic map as a map containing only drivable surface information, such as a street map or in our case a race track layout.  This is a very different representation than the sparse descriptor maps or 3-D reconstructions used for SLAM based localization.  In this work, we utilize a metric map generated by surveying the centerline of our race track.  A distance transform of this centerline results in the schematic map shown at the top of Fig. \ref{fig:pf-diagram}.  

Instead of using an existing whole image or key point based SLAM system, the monocular camera images are used as the input to a convolutional neural network in order to directly regress the free space in front of the vehicle.  Key point based localization on the AutoRally system is difficult due to the low vantage point, short stereo camera baseline, and many self-similar textures in the environment.  Our track surface is dirt and the track barriers are black, resulting in very few useful features within range of our stereo cameras suitable for key point SLAM.  Additionally, the on board cameras are subject to high amplitude vibrations during fast, aggressive driving, this breaks many SLAM algorithms that assume a low acceleration motion model.
%
%

We use a particle filter to combine the proprioceptive IMU and wheel speed sensors with the deep neural network cost map sensor.  This allows us to fully utilize all available information to drive aggressively without the aid of external localization.  In addition, we create a new neural network architecture that can be trained recurrently and utilizes temporal information to resolve difficult difficult or ambiguous visual situations.  This improves the accuracy of the cost map prediction network over the single frame case, and greatly improves performance on difficult and ambiguous cases.
%
%


The full aggressive driving system (see Fig. \ref{fig:systemDiagram}) has three main components.  First, a CNN is used to predict local cost maps from monocular input images.  Second is the particle filter.  It takes as input IMU measurements (linear accelerations and angular velocities), vehicle wheel speed measurements, and local cost map estimates from the cost map neural network.  The IMU measurements are used to propagate the state forward, and particles are weighted using wheel speed and difference between cost map predictions and our surveyed cost map.  Third is the AutoRally vehicle.  This vehicle has all of the onboard computation and sensing needed to implement this approach and is capable of high speed, aggressive driving. 



\subsection{CNN track detection module}
\label{sec:cnn}

\begin{figure}[htbp]
\begin{center}
{
\includegraphics[height=0.37\columnwidth]{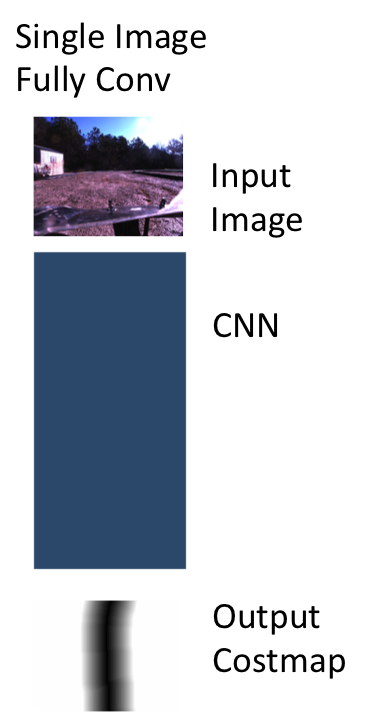}
\includegraphics[height=0.37\columnwidth]{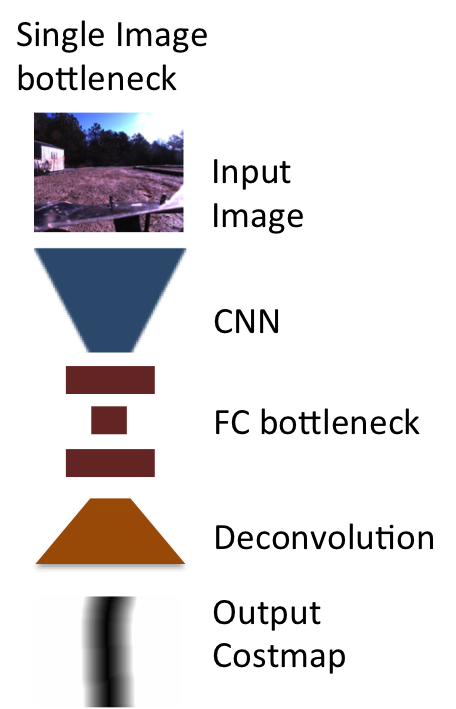}
\includegraphics[height=0.37\columnwidth]{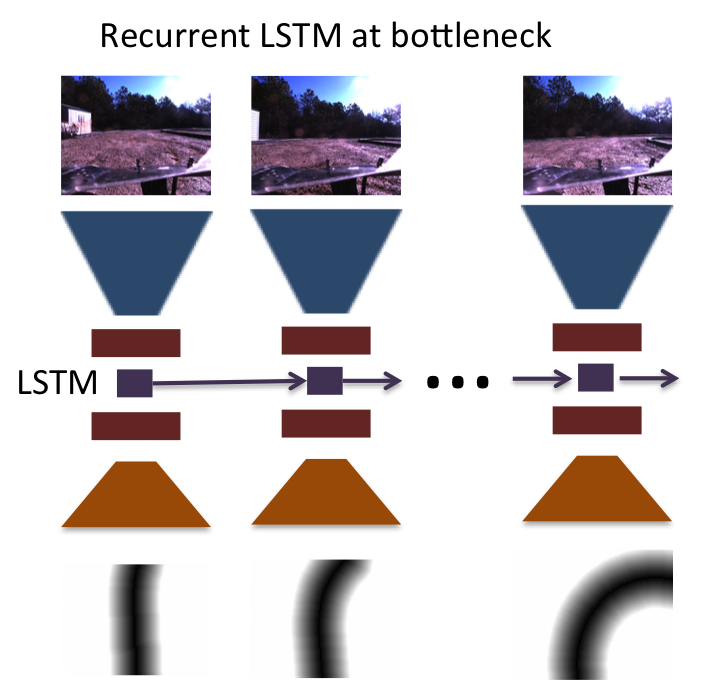}
}
\caption{Network architectures with input and training targets.  Left: Fully convolutional neural network architecture.  Dilated convolutions are used to increase receptive field.  Center: Encoder decoder neural network architecture.  Right: Encoder decoder architecture trained recurrently, with an LSTM replacing the bottleneck fully connected layer.}
\label{fig:netArchitecture}
\end{center}
\end{figure} 

We use a CNN to generate a cost map style overhead projection of the track in front of the vehicle from a monocular image stream.  Our CNN architecture is constrained to run in real time on the low power Nvidia GTX1050 GPU available on our platform, along with the model predictive controller and particle filter.  The CNN directly produces a dense cost map in the egocentric frame, with the current vehicle position at the bottom center of the image. We compare three different neural network architectures as shown in Fig. \ref{fig:netArchitecture}.

This work is an extension of our prior work in \cite{drews17aggressive}.  In that work, we used a fully convolutional neural network composed entirely of convolutions and dilated convolutions, refered to hereafter as the flat network.  In this work, we utilize an encoder-decoder architecture which improves over that network in several ways.  First, since the input image is reduced to a small hidden state, we can easily add recurrence to the network to take advantage of temporal information and resolve some cases which are ambiguous in the single frame case.  We compare the the encoder-decoder and recurrently trained encoder-decoder cases to highlight this performance improvement.  Second, because the network output is the result of deconvolution \cite{Noh2015learning}, or reverse strided convolution, the resulting cost map is much cleaner than the output of fully convolutional network.  

Third, the fully convolutional network constrains the output cost map size to match the input image size or an integer multiple of it.  However, the encoder-decoder architecture can have arbitrarily shaped output cost maps.  We use this ability to further optimize the output of the network to the particle filter.  Previously, the output of the network was set to an area of 10.6m wide by 8.5m high (160x128 at 15 pixels per meter).  However, this size is optimized for direct planning and control on the output cost map.  The particle filter uses a 5m wide by 7m high crop of this output to achieve optimal performance.  By using the ability of the encoder-decoder to output arbitrarily shaped cost maps, we can directly output a 5m x 7m image (40x56 pixels) at 8 pixels per meter.  This size was empirically determined to be optimal for the particle filter.  We find that this significantly increases the performance of the particle filter as shown in Tables \ref{tab:dataset_results_full} and \ref{tab:dataset_results_generalization}.

Using these two network architectures, we are able to maintain low latency and a frame rate of 60 Hz(full camera frame rate). Input images come directly from a PointGrey Flea3 color camera opearting at 1280x1024 resolution.  These images are downsampled to 160x128 and the images are mean subtracted and divided by the standard deviation from the training dataset.  During training, the cost map output is trained to minimize the L1 distance to the pre-computed ground truth cost maps obtained from GPS data
\begin{equation}
MAE(\theta,U_t) = \sum_{(u,v)} |I_{u,v} - \hat{I}_{u,v}(\theta,U_{t},S_{t-1})|
\end{equation}
where the L1 error $MAE$ at input image $U_t$ is a function of the CNN parameterized by $\theta$.  This CNN is also a function of the previous hidden state $S_{t-1}$.  The error is the difference between the estimated cost map $\hat{I}$ and true cost map $I$ summed over all pixels $u,v$ in the image.

L1 loss is utilized over L2 loss due to its outlier rejection.  Because it is minimizing posterior expectation of the median instead of the mean, it is less sensitive to outliers in our dataset.  However, the L1 loss is less stable during training and tends to fall into a local minima where the dataset median is predicted regardless of the input, making hyper-parameter tuning more difficult. During training, recurrent networks often required weight initialization from single-frame trained networks in order to converge.  

All networks were trained using the Adam \cite{kingma14adam} optimization algorithm in Tensorflow \cite{tensorflow2015}.  A mini-batch size of 16 images was used during training, and a small random perturbation to the white balance of each image (multiplying each channel by a normally distributed random variable between 0.9 and 1.1) was also applied. For all networks, training was stopped after the testing error plateaued.

Each image is recorded with its ground truth position and orientation using high accuracy RTK GPS position combined with IMU data.  This centimeter accurate position allows us to generate a ground truth cost map to use as a training target for each image.  These images are split into a train and test set.  Training accuracy is reported for the train set, and test set accuracy is used to determine when to stop training.  Validation accuracy is reported using the held-out data from the on-policy particle filter runs.  The networks were trained using the Adam \cite{kingma14adam} optimization algorithm in Tensorflow \cite{tensorflow2015}.  A mini-batch size of 16 images was used during trainingFor all networks, training was stopped after the testing error plateaued.

\begin{figure}[htbp]
	\begin{center}
    \includegraphics[width=0.7\columnwidth]{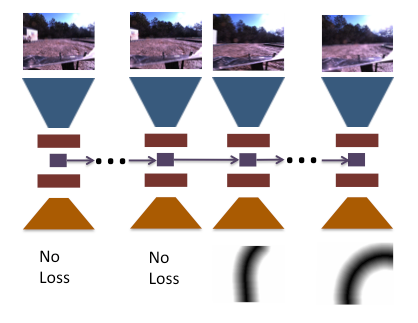}
    \end{center}
    \caption{During training, recurrent neural networks only begin accumulating loss after seeing part of the sequence of input images.}
    \label{fig:burnInTraining}
\end{figure}

In order to train the recurrent networks, the hidden state needs to be initialized.  We initialize the hidden state to zero and allow the network to burn in for some number of frames without penalizing its output as shown in Fig. \ref{fig:burnInTraining}.  We find that allowing the network to run for 8 frames without penalizing its output, and then accumulate training error for the next 8 frames produced the best results, for a total of 16 input images seen per training sample.  We found that training over longer sequences did not significantly improve testing accuracy or final particle filter performance, and made training more time consuming and more prone to getting stuck in local minima.  The network was also fairly invariant to the length of time given to burn-in.

\subsection{Particle filter}

Particle filters are a class of recursive Bayesian filters which attempt to approximate the state distribution with a set of samples (particles). They have the advantage of being able to handle non-linear dynamics and non-linear measurement models. However, in order to get good performance a large number of particles (numbering in the thousands) is often required. 

There are many variants of the particle filtering algorithm, in this work we use the sequential importance re-sampling (SIR) particle filter. The state-space for the particle filter consists of the following 5 variables: position in the map coordinate frame $(p_x, p_y)$, heading $(\psi)$, and body frame forward and lateral velocity $(v_x, v_y)$. The model predictive controller additionally uses roll information and heading derivative information, however these are passed directly from the IMU to the controller without an intermediate filtering step. The basic building blocks to the particle filtering implementation are a measurement model and a motion model:

\subsubsection{Measurement Model}

In this work, the two sensors that the AutoRally vehicle has in order to navigate are wheel speed sensors and a video stream from a monocular camera. The wheel speed sensors output a velocity estimate based on the rotation rate and diameter of the wheel, let $W$ denote the averaged velocity estimate of the two front wheels\footnote{We do not use the back wheels since they slip significantly when accelerating.}, then the wheel speed measurement model is:
\begin{equation}
p(W|\vx_j) = \frac{1}{\sqrt{2\pi\sigma^2}}\exp(-\frac{1}{2\sigma} \left(|v_x^j| - W \right)^2, ~~ \sigma = 2.5
\end{equation}
Where $\vx_j$ denotes the state of the $j_{th}$ particle and $v_x^j$ is the forward velocity. Note that the wheel speed measurements are always positive, hence the absolute value around the forward velocity.

\begin{figure}[htbp]
\centering
\includegraphics[width=0.75\columnwidth]{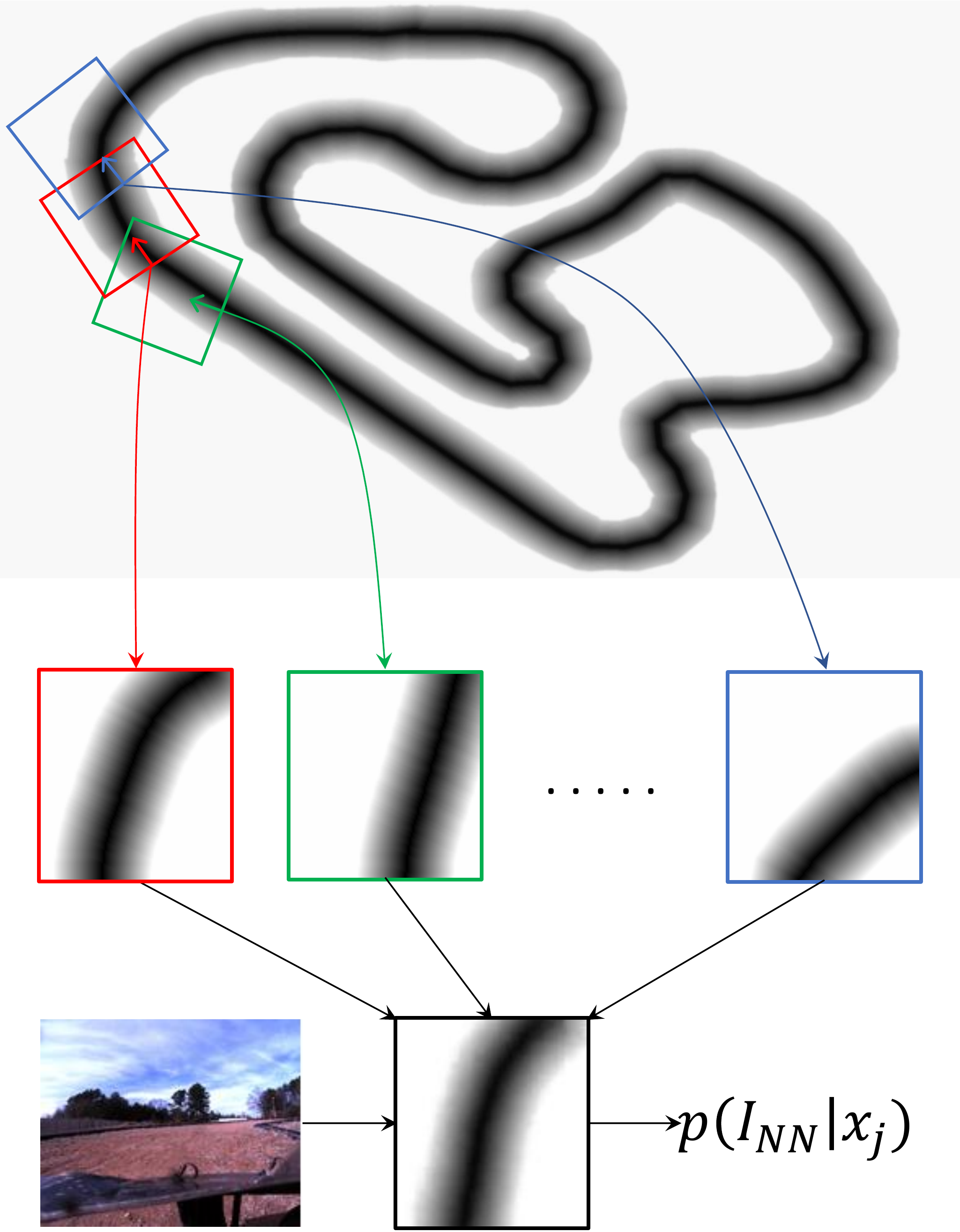}
\caption{Each particle maintains a position and heading with respect to the map coordinate frame (top), this is used to extract a top-down image of the expected local track geometry (middle), and this is compared to the output of the neural network's prediction in order to compute a measurement probability.}
\label{fig:pf-diagram}
\end{figure}

The other sensor that the AutoRally has is a monocular camera, which provides input images to a convolutional neural network, which then outputs a prediction of the local track. This prediction can be used by the particle filter to localize the vehicle in a global schematic map via the following two-step procedure:
\begin{enumerate}
\item Each particle uses its current position and heading to select the local slice of the global map corresponding to the area the neural network would be making a prediction for if the vehicle were actually in that location.
\item The local slice of the global map is compared to the actual output of the neural network, and the output of the comparison is fed into a probability distribution. The output of this distribution acts as the measurement model. 
\end{enumerate}
This procedure is visualized in Fig. \ref{fig:pf-diagram}. In this work, we use mean absolute error in order to compare the images, and feed the error into an exponential distribution. The measurement model for the neural network predictions is then:
\begin{equation}
p(I_{NN}|\vx_j) = \lambda \exp\left(-\frac{\lambda}{N}\sum_{(u,v)} \left| M_{\text{local}}^j(u,v) - I_{NN}(u,v) \right| \right)
\end{equation}
where $\vx_j$ denotes the state of the $j_{th}$ particle, $M_{\text{local}}^j$ is the associated local slice of the global map, and $I_{NN}$ is the output from the neural network. Note that $I_{NN}$ is a function of the current image and the LSTM recurrent state, which in turn is a function of all past images. The combined measurement model is then:
\begin{equation}
p(o_t | \vx_t^j) = p(I_{NN}|\vx_t^j)p(W|\vx_t^j)
\end{equation}

\subsubsection{Motion Model}
The motion model for the AutoRally is equipped with an Inertial-Measurement-Unit (IMU) which outputs accelerations $(a_x, a_y, a_z)$ and angular velocities $(\alpha_x, \alpha_y, \alpha_z)$. These are combined with standard equations for the motion of a rigid body in order to create the following noisy motion model:
\begin{align*}
\rd p_x &= \left(v_x\cos(\psi) - v_y\sin(\psi)\right) \rd t \\
\rd p_y &= \left(v_x\sin(\psi) + v_y\cos(\psi)\right) \rd t \\
\rd \psi &= \alpha_z\rd t + \sigma_\psi \rd w \\
\rd v_x &= a_x\rd t + \sigma_{v_x} \rd w, ~ \rd v_y = a_y\rd t + \sigma_{v_y} \rd w
\end{align*}
Where $\sigma_\psi = 0.275$ and $\sigma_{v_x} = \sigma_{v_y} = 0.75$. Note that this model is different than the motion model that the model predictive controller uses to control the vehicle. Future IMU measurements are obviously not available for the MPC predictions, so this model cannot be used for MPC. The model that the MPC controller uses could be used by the particle filter in place of this rigid body dynamics model, and could potentially lead to improved performance by introducing more constraints on the dynamics. However, this model provides adequate performance and is very cheap to compute. 

In our particle filter implementation, the stochastic motion model propagates forward at 200 Hz with standard Euler–Maruyama integration, measurements are processed at 20 Hz, and particle re-sampling occurs at 5 Hz. All of the motion and measurement models are implemented as CUDA Kernels on the GPU, which is necessary since every measurement update requires computing the difference between two images (the local image patches are 35x25 pixels). It is possible to run both the measurement and re-sampling loop at faster rates, however doing so did not lead to improved performance and other processes (the controller and neural network) also require GPU resources. We used 6400 particles, and computed the final state estimate as the mean of the particles.

\section{AutoRally Platform and MPPI Controller}

\subsection{AutoRally}

In order to test the performance of these algorithms in a real racing scenario, we utilize the AutoRally \cite{goldfain2018autorally} platform.  This robot is based on a 1:5 scale RC chassis capable of speeds of nearly 60mph.  It has a desktop-class Intel i7 processor and NVidia Gtx 1050 GPU for processing.  IMU, GPS, and wheel speed sensors are used, as well as images captured from an on-board Point Grey camera.  This allows all computation to be performed in real time on-board the vehicle.  All software runs under the Robot Operating System (ROS).  Particle filter and MPPI code is written in CUDA, and CNN forward inference is done using a custom ROS wrapper around TensorFlow \cite{tensorflow2015}.

\subsection{MPPI}

Model Predictive Path Integral Control (MPPI) is a stochastic model predictive control (MPC) method designed to work with non-linear dynamics, and non-convex cost objectives. It has been shown to work well in practice applied to AutoRally platform up to and beyond the friction limits of the vehicle \cite{williams2017information}.

MPPI works by quickly sampling and evaluating thousands of control sequences, and then computes the control input as a cost weighted average over the sampled controls. In order to evaluate a control sequences, a dynamics model is propagated forward in state space using the system dynamics, and each trajectory is evaluated according to a cost function. As in \cite{williams2017information}, we use a neural network model to learn the dynamics. Real time execution of MPPI on AutoRally is enabled by the onboard Nvidia GPU.

We only use a running cost function in this paper (no terminal cost). The running cost that we use for generating driving behaviors from MPPI has the form:
\begin{equation}
q(x) = w \cdot \left(C_M(p_x, p_y), h(v_x, v_x^d), 0.9^t I, \left(\frac{v_y}{v_x}\right)^2 \right),
\label{equ:costFunction}
\end{equation}
In these equations $w$ is a vector of weights.

The first cost term, $C_M(p_x, p_y)$, is the positional cost of being at the position $(p_x, p_y)$. This positional cost is obtained from the cost map when the map is in use, and directly from the output of the neural network in the case of mapless driving. The second term is a cost for achieving a desired speed $v_x^d$, the function $h$ denotes the metric used in the cost computation. There are two different modes of driving that we use in our experiments. The first is slow/medium speed driving where the speed target actually describes the speed we want the vehicle to achieve, in this case $h$ is the squared difference. The second mode is high speed driving, where the speed target is set to a value above what is physically possible for the vehicle to obtain (25 m/s in our case). In the second mode we use the absolute value instead of the squared difference. This was done because the absolute value has a constant gradient magnitude (wherever it exists), which enables the target speed to be set arbitrarily high without creating an exploding gradient problem. Note that even though MPPI is a gradient free algorithm, it is still sensitive to gradient magnitudes since trajectories are weighted relative to one another. The third term in the cost is a time-decaying indicator variable which is turned on if the track-cost, roll angle, or heading velocity are too high. The final term in the cost is a penalty on the slip angle of the vehicle.

\section{Experimental results}

\begin{figure}
\centering
\includegraphics[width=0.85\columnwidth]{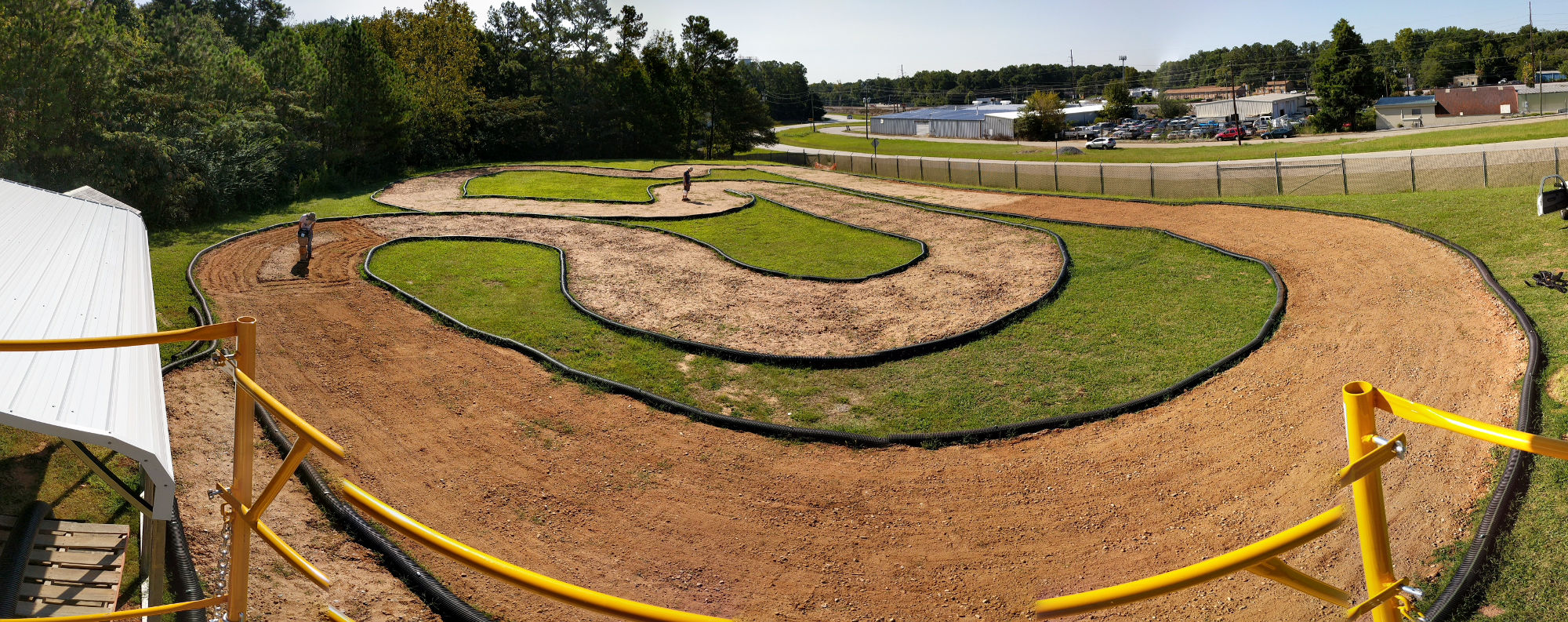}
\caption{Test track for physical vehicle experiments.  This track includes a variety of turns, and is very challenging for the visual navigation system.}
\label{fig:gtarf}
\end{figure}

All experimental results are collected using our 1:5 scale AutoRally vehicle at the Georgia Tech Autonomous Racing Facility track 2, shown in Fig. \ref{fig:gtarf}.  This challenging track includes turns of varying radius including a 180 degree hairpin and S curve, and a long straight section.  All results presented in this paper are the vehicle driving autonomously using the camera for localization.  The monocular color camera, IMU, and wheel speed sensors are the only sensors used except for the direct CNN map usage case, where vehicle velocities are derived from GPS.

\subsection{CNN training}

\begin{table}[htbp]
    \caption{Average L1 pixel accuracy}
    \begin{center}
    \begin{tabular}{|l|l|l|l|l|l|}
        \hline
         & Flat & ED & ED-R & ED-S & ED-R-S \\ \hline
        Train & 94.66 & 93.66 & 96.48 & 92.69 & 96.09 \\ 
        Val & 92.29 & 91.64 & 92.54 & 88.13 & 89.64 \\ \hline
        \hline
        \multicolumn{6}{l}{ED: Encoder Decoder. R: Recurrent. S: 40x56 output} \\
    \end{tabular}
    \end{center}
    \label{tab:l1-results}
\end{table}

We compare our recurrent encoder decoder architecture to two different single frame networks using the training objective (L1 pixel error), dataset particle filter recovered position error, and on-policy driving performance using the MPPI controller.  We compare the performance for both the fully trained and leave-one-direction-out case. 

Because the recurrent architecture was trained over sequences of data, the network was able to learn several interesting temporal patterns in the input data.  It learned to smooth the cost map prediction from frame to frame, so output does not jitter as much as the single frame network.  It also learns to integrate information over multiple frames in areas where it is difficult to see where the track goes, such as Fig. \ref{fig:lstmsequence}.  This allows the LSTM to produce much better results in difficult or ambiguous areas that cannot be easily interpreted in a single image.

Training data was gathered during the course of normal vehicle testing at the Georgia Tech Autonomous Racing Facility over the course of approximately one year, shown in Fig. \ref{fig:gtarf}.  Because this system has the ability to learn from any data, on or off policy, we utilize data collected from various experiments.  In total, we collect approximately 90k samples.  These samples are broken into about 75k training samples and 15k testing samples.  The testing samples are taken as full held-out contiguous runs to allow recurrent network testing.  In addition to the testing samples, we report validation error on a corpus of data that includes 5 test runs performed during experimentation for this paper and not used in either the training or testing sets.

All samples are labeled with ground truth position and orientation from our RTK-GPS and IMU state estimation, which is typically accurate to several centimeters.  Because the GPS is not completely reliable, part of the process of collecting training and testing samples is ensuring GPS based pose is sufficiently accurate during a run for use as ground truth.  This ground truth pose is used to automatically label the images with local cost maps from a pre-surveyed track map.

For training, testing, and validation accuracy, we report the average per-pixel frame accuracy as
\begin{equation}
A_t = 1 - \sum_{(u,v)} \frac{|I_{u,v} - \widehat{I}_{u,v}|}{N}
\end{equation}
where $N$ is the number of pixels per image and $I_{u,v}$ is a normalized pixel value in the range $0-1$.
For accuracy over sequences, each neural network computes an output per frame (with the hidden state propagated through the entire sequence for recurrent networks), and final accuracy reported as the mean per-frame accuracy.

Average training and validation accuracy for the networks is shown in Table  \ref{tab:l1-results}.  Since there is no straightforward way to normalize these results for map scale and the number of pixels available, the accuracy is not directly comparable between three large networks and the small (40x56 output) networks.  The LSTM networks achieve significantly better validation error due to learning to integrate information temporally before producing a result.  It is apparent from Fig. \ref{fig:lstmsequence} that this is at least partially due to the networks ability to integrate information over time and correctly identify visually challenging frames.  Single frame networks are not able to identify the track in these challenging cases.

\subsection{Particle Filter}

Localization performance of the particle filter, as well as real-world performance of the full system, was measured by driving the vehicle at a 6 m/s target speed for 3 laps using each of 5 neural networks (approximately 100s of driving for each condition).  These 5 networks are trained in 2 different ways.  First, using all available data.  This includes clockwise and counterclockwise training data from 7 days of testing.  Second, we perform leave-one-direction-out (lodo) testing by training neural networks on only the subset of data where the vehicle is traveling clockwise.  All testing and validation is performed with the vehicle traveling counterclockwise.  Average position error is reported for both the on-policy case where the vehicle is driven using the pose estimate and the off policy case where the filter is run off-line on recorded data and the error is calculated.


\subsubsection{Full Dataset Training}

\begin{figure*}[htbp]
    \begin{center}
    {
    \includegraphics[height=0.5\columnwidth]{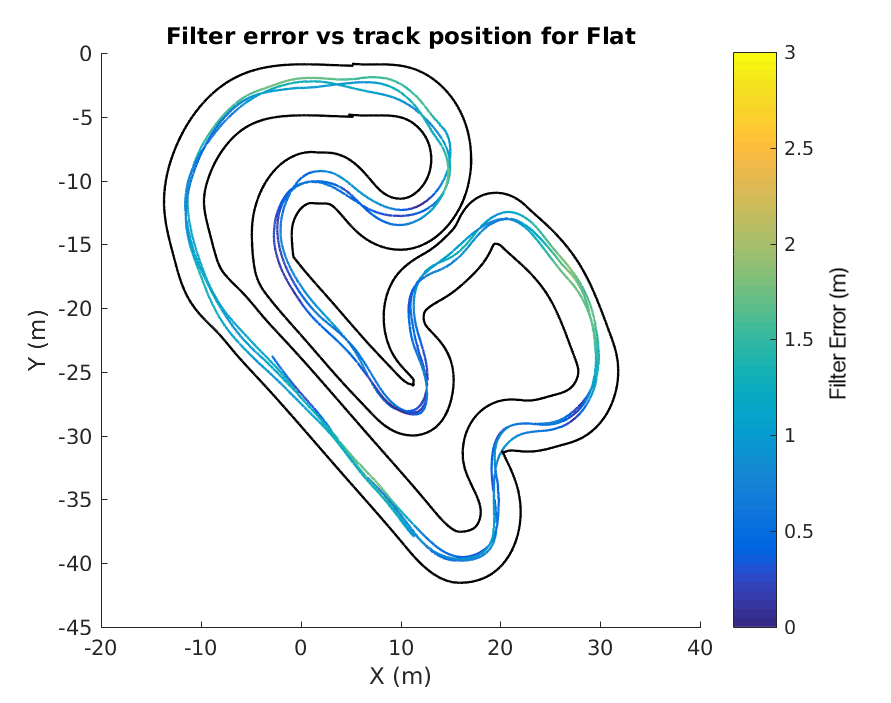}
    \includegraphics[height=0.5\columnwidth]{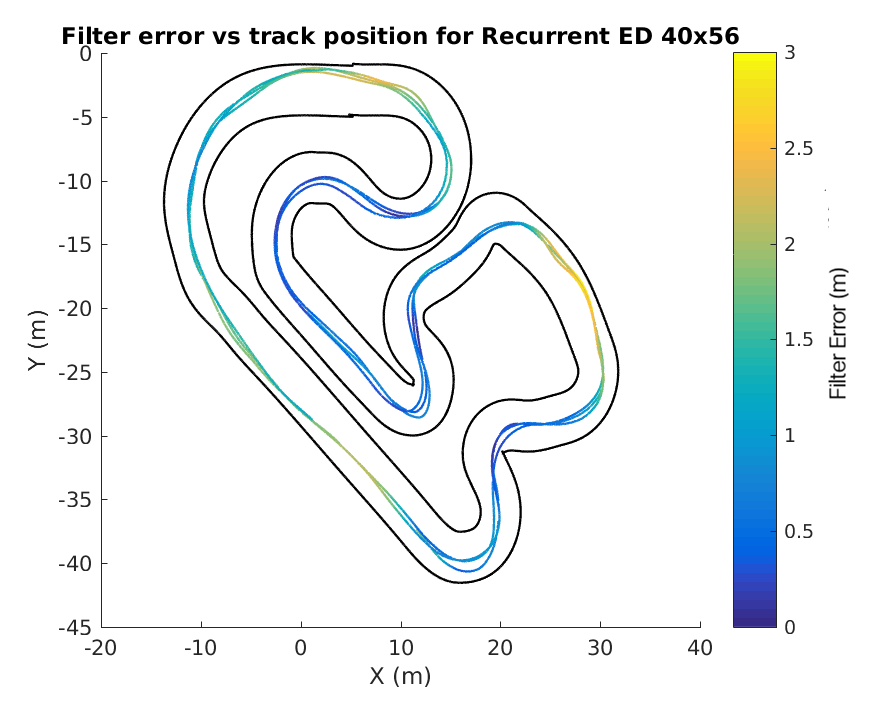}
	\includegraphics[height=0.5\columnwidth]{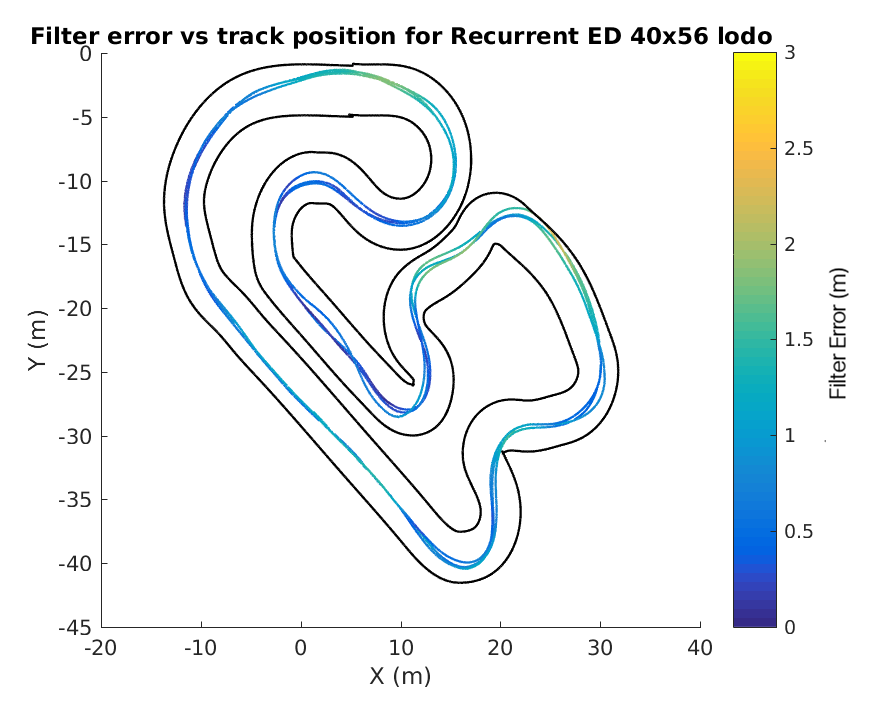}

    }
    \caption{Particle filter position estimate error plotted vs ground truth position.  For each network, MPPI was run with a target speed of 6 m/s using the particle filter pose estimate as its pose source.  MPPI planned using position, velocity, and orientation from the particle filter and track cost from the pre-surveyed map. }
    \label{fig:accuracy_position_plots}
    \end{center}
    \vskip -0.2in
\end{figure*}

\begin{table}[htbp]
\caption{Particle Filter Position Error for Full Dataset Trained Neural Networks}

\begin{center}
\begin{tabular}{l|ccccc} Dataset$\downarrow$ & ED & ED-S & Flat & ED-R & ED-R-S \\ \hline 
    ED & \cellcolor{blue!25}1.21 & 1.12 & 1.01 & 1.19 & 1.0\\ 
    ED-S & 1.27 & \cellcolor{blue!25}1.19 & 1.02 & 1.11 & 1.07\\ 
    Flat & 1.04 & 0.937 & \cellcolor{blue!25}0.918 & 0.979 & 0.894\\ 
    ED-R & 1.08 & 0.985 & 1.05 & \cellcolor{blue!25}1.01 & 1.05\\ 
    ED-R-S & 1.17 & 1.01 & 0.879 & 1.09 & \cellcolor{blue!25}1.09\\ 
    ED-lodo & 1.22 & 0.91 & 1.45 & 1.06 & 1.05 \\
    ED-S-lodo & 0.92 & 0.86 & 0.87 & 0.88 & 0.84 \\
    ED-R-S-lodo & 0.93 & 0.96 & 0.81 & 1.06 & 0.80 \\
    \hline 
    Average & 1.11 & 1.00 & 1.00 & 1.05 & 0.97\\
\end{tabular}
\end{center}
\label{tab:dataset_results_full}
\end{table}

In the full training case, all network architectures were able to successfully drive the vehicle in the on-policy condition.  The network errors are summarized in Table \ref{tab:dataset_results_full}.  The small recurrent encoder-decoder achieved the best average position error when tested against recorded data for all 8 3-lap on policy runs.  It significantly out-performs the non-recurrent encoder-decoder, demonstrating the benefits of recurrence.  Figure \ref{fig:accuracy_position_plots} shows the position error as a function of ground truth track position.  We can see that the error for the recurrent network tends to be concentrated in shorter sections of the track, and that there are some track sections that are difficult for all networks.

\subsubsection{Leave One Direction Out}

\begin{table}[htbp]
\caption{Particle Filter Position Error using Leave One Direction Out Neural Networks}
\begin{center}
\begin{tabular}{l|ccccc} Dataset$\downarrow$ & ED & ED-S & Flat & ED-R & ED-R-S\\ \hline 
    ED & 1.18 & 1.03 & 1.06 & 0.954 & 1.04\\ 
    ED-S & 1.15 & 1.03 & NA & 0.779 & 0.99\\ 
    Flat & 0.981 & 0.904 & 0.812 & NA & 0.854\\ 
    ED-R & 1.11 & 1.12 & 0.91 & 0.823 & 0.986\\ 
    ED-R-S & NA & NA & NA & 1.08 & 1.19\\ 
    ED-lodo & \cellcolor{blue!25}1.05 & 1.02 & 1.23 & 1.46 & 1.04\\
    ED-S-lodo & 0.96 & \cellcolor{blue!25}0.90 & 1.11 & 0.61 & 0.84 \\
    
    ED-R-S-lodo & 0.93 & 0.91 & 0.97 & 0.62 & \cellcolor{blue!25}0.85 \\
    \hline
    Average & 1.05 & 0.99 & 1.02 & 0.90 & 0.97 \\
    \multicolumn{6}{c}{ED: Encoder Decoder. R: Recurrent. S: 40x56 output.} \\
    \multicolumn{3}{r}{NA: failed to initialize} & \multicolumn{2}{c}{\cellcolor{blue!25}on policy} & \\
\end{tabular}
\end{center}
\label{tab:dataset_results_generalization}
\end{table}

We find, in the leave-one-direction-out case, that the encoder-decoder and recurrent encoder-decoder networks are better able to generalize to this new environment and maintain a sufficiently accurate pose than the flat network, as demonstrated in Table \ref{tab:dataset_results_generalization}.  In the on-policy case, the flat network and the large recurrent encoder-decoder were unable to maintain an accurate pose, and caused the vehicle to crash into a barrier before completing 3 laps.  Both small output networks and the large non-recurrent encoder-decoder were able to successfully drive the vehicle at a 6m/s target speed for 3 laps.  In the off-policy case, the particle filter using the flat network failed to initially converge in 2 of 5 cases.  The small recurrent encoder-decoder successfully initialized the particle filter and tracked pose through all datasets, with the encoder decoder networks achieving lower overall error than the flat network.


\subsection{Direct Driving}

\begin{table}[htbp]
    \caption{Lap times for Direct Driving and Particle Filter at 6 m/s target speed}
    \begin{center}
    \begin{tabular}{|l|l|l|l|l|l|}
        \hline
        Method & Flat & ED & ED-R & ED-S & ED-R-S \\ \hline
        Direct & 37.3 & 37.8 & 39.1 & NA & NA \\ \hline
        Particle Filter & 35.6 & 34.8 & 35.9 & 34.5 & 35.3 \\ \hline
        \multicolumn{6}{c}{ED: Encoder Decoder. R: Recurrent. S: 40x56 output} \\
    \end{tabular}
    \end{center}
    \label{tab:dd_times}
\end{table}

In order to test these methods against previous work, we allow MPPI to plan directly on the output of each image as described in \cite{drews17aggressive}.  Both small networks, that output 5m x 7m regions, were unstable using this method and crashed almost immediately.  This is due to the output image size being too skinny for the network to plan a reasonable path on.  The summary of lap times for the other methods are show in Table \ref{tab:dd_times}.  However, due to being able to plan for a longer time horizon, the particle filter method achieved shorter lap times over all tested architectures.

\subsection{Performance limits}

\begin{figure}[htbp]
\centering
\includegraphics[width=0.8\columnwidth]{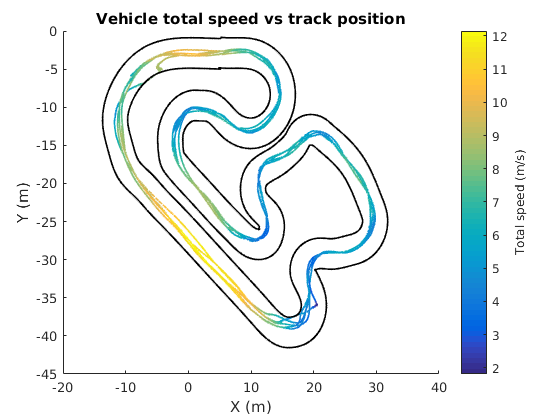}
\includegraphics[width=0.8\columnwidth]{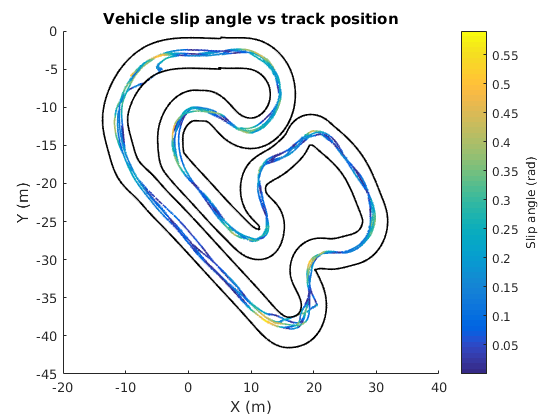}
\label{fig:fastRun}
\caption{Vehicle speed and slip angle and speed as it traverses 5 laps at the limits of performance.  The vehicle needs to be able to drive from 2 m/s to 12 m/s, at slip angles up to 32 degrees, in order to maximize lap times.}
\end{figure}

In order to test the limits of our localization system, we tasked MPPI with traversing the track as quickly as possible. Using this method, we set a lap time of \textbf{27.9s at 12.2 m/s (27.3 mph) top speed}.  Figure \ref{fig:fastRun} shows that the algorithm is able to reliably perform 5 laps while sustaining high speeds and high slip angles.  For comparison, our previous best published lap time at this track is 32s at 8.5 m/s maximum speed, and the best lap in all training data is 29.4s at 10.4 m/s.  This method is able to consistently push the limits of the vehicle and execute aggressive maneuvers using only a monocular camera, IMU, and wheel speed sensors.

\section{Conclusion}

In this work, we demonstrate a system capable of repeatable aggressive driving on a complex dirt test track using only monocular vision, IMU, and wheel speed measurements.  It combines neural network cost map regression from a monocular camera, particle filter based state estimation, and model predictive control running in real time on a rugged, high speed autonomous system.  We demonstrate CNN system performance improvement using LSTMs to learn to integrate temporal information.  We demonstrate the ability of a particle filter to integrate this information with IMU and wheel speed sensors and produce a high-rate, high accuracy state estimate.  We demonstrate the ability of our encoder-decoder network to generalize to traversing the track in a direction it has not seen before by performing leave on direction out experiments on both datasets and on the physical system.  Finally, we demonstrate the full system performing at the limits of handling of our platform by executing high speed laps at our test track using only monocular vision for position information.

This work demonstrates the advantages of combining model predictive control with state estimation and learned perception.  Where states are well understood, in the case of vehicle pose estimation, we use traditional pose estimation and model predictive control solutions (with the difficult estimation problem, vehicle dynamics prediction from control inputs, still being learned).  The perception system still contains useful state information, but the exact structure of this state is less well understood.  Therefore, an LSTM neural network model is the correct tool to model these video input images.


As with any real-world system, we find that details significantly improve the performance of the system.  Reducing system lag by A) propagating state forward at 200 Hz using the IMU measurements, B) threading software systems where possible to reduce latency and C) forward propagating the model predictive control outputs using state feedback gains all help to push the system to its limit.  When training our neural network, careful curation of the dataset, model architecture tuning, and hyper-parameter tuning all improve final system performance.  Careful system identification dataset collection improves the accuracy and predictive power of the vehicle dynamics model.  Finally, having a system that is robust and able to be repeatedly pushed to the limits of handling (and beyond) is crucial to iterative design process needed to push the system to the limits of handling and grip.

\bibliographystyle{IEEEtran}
\bibliography{path_integral_bib,deepnet}

\end{document}